%% file: root.tex
\documentclass[letterpaper, 10 pt, conference]{IEEEtran}  
\pdfoutput=1

\IEEEoverridecommandlockouts                              

\usepackage[utf8]{inputenc}
\usepackage[T1]{fontenc}
\usepackage{graphicx}
\usepackage[section]{placeins}
\usepackage{multirow}
\usepackage{booktabs}
\usepackage{amsmath}
\usepackage{amsfonts}
\usepackage{amssymb}
\usepackage{csquotes}

\usepackage{blindtext}  
\usepackage[backend=bibtex,style=numeric,sorting=none]{biblatex} 

\usepackage{tikz}

\title{\LARGE \bf Continuous Decoding of Daily-Life Hand Movements from Forearm Muscle Activity for Enhanced Myoelectric Control of Hand Prostheses*}

\author{Alessandro Salatiello$^{1,2}$ and Martin A. Giese$^{1}$, {\it Member, IEEE} 
\thanks{*This work was supported by: BMBF FKZ 01GQ1704, HFSP RGP0036/2016, KONSENS-NHE BW Stiftung NEU007/1, DFG GZ: KA 1258/15-1, and ERC 2019-SyG-RELEVANCE-856495.}
\thanks{$^{1}$A. Salatiello and M. A. Giese are with the Section for Computational Sensomotorics, Department of Cognitive Neurology, Centre for Integrative Neuroscience, Hertie Institute for Clinical Brain Research, University Clinic Tübingen, Otfried-Müller-Straße 25, 72076 Tübingen, Germany {\tt\small\{alessandro.salatiello, martin.giese\}@uni-tuebingen.de}}
\thanks{$^{2}$A. Salatiello is with the International Max Planck Research School for Intelligent Systems, Max-Planck-Ring 4, 72076 Tübingen, Germany.} 
}

\newcommand\copyrighttext{%
  \footnotesize \textcopyright 2021 IEEE. Personal use of this material is permitted.
  Permission from IEEE must be obtained for all other uses, in any current or future
  media, including reprinting/republishing this material for advertising or promotional
  purposes, creating new collective works, for resale or redistribution to servers or
  lists, or reuse of any copyrighted component of this work in other works.}
\newcommand\copyrightnotice{%
\begin{tikzpicture}[remember picture,overlay]
\node[anchor=south,yshift=10pt] at (current page.south) {\fbox{\parbox{\dimexpr\textwidth-\fboxsep-\fboxrule\relax}{\copyrighttext}}};
\end{tikzpicture}%
} 
\begin{document}
\renewcommand{\thefootnote}{\roman{footnote}}

\maketitle
\copyrightnotice


\thispagestyle{empty}
\pagestyle{empty}

\begin{abstract}
\input{Abstract}
\end{abstract}

\section{INTRODUCTION}
\input{Introduction}

\section{METHODS}
\input{Methods}

\section{RESULTS}
\input{Results}

\section{DISCUSSION}
\input{Conclusion}





\printbibliography

\end{document}

%% file: Abstract.tex
State-of-the-art motorized hand prostheses are endowed with actuators able to provide independent and proportional control of as many as six degrees of freedom (DOFs). The control signals are derived from residual electromyographic (EMG) activity, recorded concurrently from relevant forearm muscles. Nevertheless, the functional mapping between forearm EMG activity and hand kinematics is only known with limited accuracy. Therefore, no robust method exists for the reliable computation of control signals for the independent and proportional actuation of more than two DOFs.
A common approach to deal with this limitation is to pre-program the prostheses for the execution of a restricted number of behaviors (e.g., pinching, grasping, and wrist rotation) that are activated by the detection of specific EMG activation patterns. However, this approach severely limits the range of activities users can perform with the prostheses during their daily living. 
In this work, we introduce a novel method — based on a long short-term memory (LSTM) network — to continuously map forearm EMG activity onto hand kinematics.  Critically, unlike previous work, which often focuses on simple and highly controlled motor tasks, we tested our method on a dataset of activities of daily living (ADLs): the KIN-MUS UJI dataset. To the best of our knowledge, ours is the first reported work on the prediction of hand kinematics that uses this challenging dataset. Remarkably, we show that our network is able to generalize to novel untrained ADLs. Our results suggest that the presented method is suitable for the generation of control signals for the independent and proportional actuation of the multiple DOFs of state-of-the-art hand prostheses.

%% file: Introduction.tex
The high costs and the often poor functional outcomes associated with hand transplantation \cite{aman2019bionic} have spurred considerable research efforts into designing hand prostheses. Significant progress in mechatronics has allowed the development and commercialization of motorized hand prostheses\footnotemark\textsuperscript{,}\footnotemark\textsuperscript{,}\footnotemark\textsuperscript{ } endowed with as many as eighth degrees of freedom (DOF) actuated by up to six motors \cite{calado2019review}. 
Despite these technological advances, major limitations in {\it neural interfacing} strongly affect the usability and functionality of these devices \cite{farina2014extraction}. Such limitations are generally so severe that prosthetic hands often have functional capabilities not dissimilar to transplanted hands, tend to be considered as mere assisting tools by their users \cite{salminger2016functional}, and have an average abandonment rate as high as 23\% \cite{biddiss2007upper}.\\
\indent Both invasive and non-invasive interfaces with either the central \cite{collinger2013high,meng2016noninvasive} or the peripheral \cite{wendelken2017restoration,hahne2018simultaneous} nervous system are theoretically suitable to extract adequate control signals for the prostheses.
However, in practice, non-invasive interfaces with the peripheral system — that record surface electromyographic signals (EMG) from remnant muscles \cite{grushko2020control} — are often preferred in clinical settings due to their better cost-effectiveness.
Despite being non-invasive, such methods still allow the extraction of information about the neural drive to the muscles, and thus about the motor task the user wants to perform \cite{farina2014extraction}. Whenever possible, EMG signals are recorded from the forearm, where the 15 extrinsic muscles mediating wrist and finger movements are located \cite{okwumabua2020anatomy}. However, despite the large body of research aimed at understanding the {\it functional mapping} between forearm EMG activity patterns and hand movements \cite{long1964electromyographic,schieber1995muscular,li2000contribution}, this mapping is still currently known only with limited accuracy and is thus the subject of an active area of investigation \cite{beringer2020effect}. Consequently, up to date, there is no established method to extract reliable control signals from the EMGs to actuate all the DOFs of state-of-the-art prostheses independently and proportionally. \input{activitiesTable} \\
\indent A traditional strategy to overcome this obstacle is to program the prostheses to only actuate a single DOF at a time. Such a strategy ensures that the extraction of two distinct activation patterns (one per direction) suffices to have robust control over the selected DOF. However, in this case, users have to use time-consuming heuristics (e.g., co-contractions) to switch between DOFs, until the desired one is selected \cite{ciancio2016control}. Most modern hand prostheses adopt a similar principle: they can still only recognize two activation patterns, but these are directly mapped to distinct hand functions (e.g., grip types) rather than individual joint movements \cite{calado2019review}. 

Recent research efforts have led to the development of techniques based on {\it pattern recognition} that can detect up to 12 activation patterns \cite{naik2015transradial,vidovic2015improving,yun2017maestro}. This strategy (implemented also in recent commercial devices\footnotemark\textsuperscript{,}\footnotemark) eliminates the need to issue specific commands to switch between functions. 
\addtocounter{footnote}{-4}
\footnotetext{https://tinyurl.com/ottobockus-bebionic}\stepcounter{footnote}
\footnotetext{https://tinyurl.com/ossur-i-limb}\stepcounter{footnote}
\footnotetext{https://www.taskaprosthetics.com/}\stepcounter{footnote}
\footnotetext{https://coaptengineering.com/}\stepcounter{footnote}
\footnotetext{https://www.i-biomed.com/}

Despite the increase in efficiency, pattern recognition approaches still suffer from significant limitations. First, the users can only interact with the environment using the limited set of actions the prostheses can execute. Second, these actions can only be executed one at a time; this implies that, for example, the users cannot simultaneously perform a grasp and a wrist rotation (e.g., to rotate a doorknob) unless this combined action is explicitly programmed. Third, the users typically have little or no control over action speed and applied force. 

{\it Regression-based} methods promise to address some of these issues by allowing independent, simultaneous and proportional control of multiple DOFs, and have been attracting the attention of several research groups \cite{hahne2014linear,hwang2017real}. Such methods try to map EMG signals — concurrently recorded from multiple sites — onto the movement velocity of the available DOFs. A few recent studies conducted with amputees have demonstrated the robustness of linear regression methods and their superior functional performance compared to conventional control techniques in both laboratory settings \cite{hahne2018simultaneous} and daily life \cite{hahne2020longitudinal}. Nevertheless, the number of controlled DOFs in these studies was as small as two (namely, wrist rotation and hand aperture), which significantly limited participants' dexterity. For instance, they could still not control their prosthetic fingers individually, which is one of the most desired features among amputees \cite{cordella2016literature}.

Numerous research groups have investigated the use of non-linear regression methods to extract richer control signals from the forearm EMGs for the independent and proportional control of multiple DOFs (e.g., \cite{ngeo2014continuous,ghaderi2015hand,krasoulis2015evaluation,zhang2020simultaneous}). These studies achieved remarkable offline prediction of the kinematics of up to 22 hand joint angles by using simultaneously recorded multisite forearm EMG activation patterns. Nonetheless, the simple, repetitive, and highly controlled motor behaviors used in these studies cast doubt on the ability of the proposed methods to generalize to realistic motor behaviors and online settings. As a matter of fact, it is well documented that the EMG patterns recorded from movement-relevant muscles during the execution of specific movements involving few joints might change drastically depending on the configuration of other still joints. For example, \cite{beringer2020effect} reported wrist-position-dependent amplitude modulations of finger flexor EMGs up to 70\% during the execution of individual finger movements. This might partially explain the poor correlations observed between offline and online myoelectric control performance \cite{hwang2017real}.

These reasons motivate the need to develop and validate prediction algorithms using EMGs and kinematics recorded during the execution of realistic motor behaviors. To this end, in this work, we introduce a long short-term memory (LSTM) network \cite{hochreiter1997long} to perform the online mapping of the forearm EMG activities, recorded during the execution of everyday activities, onto the corresponding hand kinematics. Specifically, we test our approach on a recently published dataset of activities of daily living (ADLs), the KIN-MUS UJI dataset \cite{jarque2019calibrated}. Remarkably, we show that the trained network is able to generalize to novel ADLs that were not used during training. To the best of our knowledge, ours is the first reported work on the prediction of this dataset.

%% file: activitiesTable.tex
 \aboverulesep=0ex
 \belowrulesep=0ex
 \renewcommand{\arraystretch}{1.1}
\begin{table*}[t]
\caption{Activities of daily living used for training, validating, and testing the networks}
\label{tab:my-table}
\begin{tabular}{@{}ll|ll|ll@{}}
\toprule
\multicolumn{2}{c|}{Training Set}                         & \multicolumn{2}{c|}{Validation Set} & \multicolumn{2}{c}{Test Set}              \\ \midrule
ID & Description                                          & ID  & Description                   & ID & Description                          \\ \midrule
1 & Coin: from table to change purse & 21 & Milk carton: pour content into jug & 24 & Toothpaste tube: squeeze content on toothbrush \\
2  & Zip: open and close                                  & 22  & Jug: pour content into glass  & 25 & Spray bottle: spray content on table \\
3 & Coin: from change purse to table & 23 & Glass: pour content into jug       & 26 & Cloth: wipe table (5 sec)                      \\
4  & Wooden cube: pick and place (twice)                  &     &                               &    &                                      \\
5  & Iron: pick and place                                 &     &                               &    &                                      \\
6  & Screwdriver: grasp and screw 360$^{\circ}$           &     &                               &    &                                      \\
7  & Nut: grasp and screw onto bolt                       &     &                               &    &                                      \\
8  & Key: grasp, insert into lock, and turn 180$^{\circ}$ &     &                               &    &                                      \\
9  & Door handle: turn 30$^{\circ}$                       &     &                               &    &                                      \\
10 & Shoelaces: tie                                       &     &                               &    &                                      \\
11 & Jar lid: grasp, unscrew, and place (twice)           &     &                               &    &                                      \\
12 & Button: pass through buttonholes (twice)             &     &                               &    &                                      \\
13 & Compression stockings: apply to left arm             &     &                               &    &                                      \\
14 & Knife: cut piece of clay                             &     &                               &    &                                      \\
15 & Spoon: bring to mouth (five times)                   &     &                               &    &                                      \\
16 & Pen: pick, write own name, and place                 &     &                               &    &                                      \\
17 & Paper sheet: fold and insert into envelope           &     &                               &    &                                      \\
18 & Paper clip: put on envelope                          &     &                               &    &                                      \\
19 & Keyboard: type                                       &     &                               &    &                                      \\
20 & Phone: grasp, bring to ear, and place                &     &                               &    &                                      \\ \bottomrule
\end{tabular}
\end{table*}

%% file: Methods.tex
\begin{figure*}[ht]
 \center
    \includegraphics[width=\textwidth]{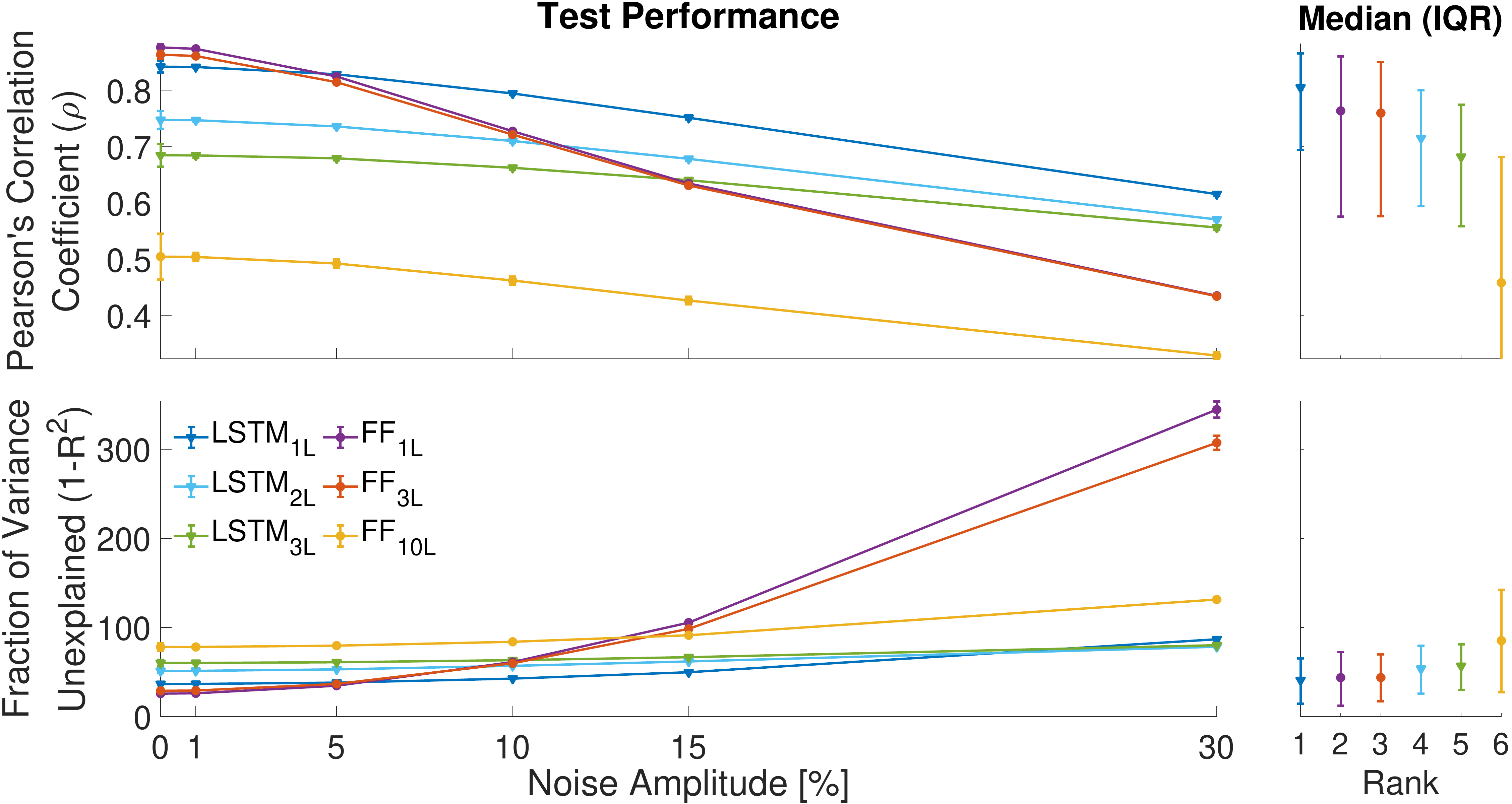}
  \caption{\textbf{Robustness analysis.}  (Top): Pearson's correlations between ground-truth and predicted angular accelerations for all the trained networks. The results are averaged across joints, tasks and noise realizations and reported as a function of noise amplitude. (Bottom): corresponding fraction of ground-truth data variance unexplained by the predictions. (Right): medians and interquartile ranges (IQR) of performance measures pooled across joints, tasks, noise levels, and noise realizations; medians are ranked in order of decreasing performance.}
    \label{fig1}
\end{figure*}

\subsection{Dataset}
To validate our method, we tried to learn a stable functional mapping between the forearm EMG activities and hand kinematics recorded during the execution of activities of daily living (ADLs). To this end, we used the KIN-MUS UJI dataset \cite{jarque2019calibrated}. This dataset contains recordings of muscle activities (7 channels) and hand kinematics (18 DOFs) of 22 participants during the execution of 26 representative activities of daily living. Each activity was performed by the participants only once and lasted several seconds. Further details about the performed activities are reported for convenience in Table \ref{tab:my-table}.

The muscle activities were recorded using surface bipolar electrodes, whose locations were chosen to maximize the extraction of information generated by the forearm muscles \cite{jarque2018identification}. The hand kinematics were recorded using an instrumented glove (CyberGlove, CyberGlove Systems LLC), which tracked fingers' and wrist's joint angles. The recorded ADLs include the 20 actions of the Sollerman Hand Function Test (SHFT) \cite{sollerman1995sollerman}, which are commonly used to assess hand function in clinical settings and involve the interaction with objects of different sizes and weights, such as coins, cutlery, and irons. All the analyses reported in this work are based on the data from participants 1 through 20. Data from participants 21 and 22 presented missing data and were thus discarded.

Importantly, most proposed methods for the online mapping of forearm EMGs onto hand kinematics and for the classification of hand movements are validated using the Ninapro database \cite{atzori2014electromyography}. However, the datasets included in this database mainly consist of simple and highly controlled movements, such as fingers' flexion with static wrist posture and different types of grips. This is in stark contrast with the dataset we used in this work, where every task involves a markedly different form of interaction with daily living objects, in addition to a reaching and a release phase. For this reason, mappings learned on this dataset promise a superior ability to generalize to online settings.
\subsection{Signal Preprocessing}
The hand joint angles were acquired at a sampling rate of 100Hz. The EMG signals were acquired at a sampling rate of 1000Hz with a passband between 20Hz and 460Hz. The kinematic data were then zero-phase lowpass filtered with a Butterworth filter (cut-off frequency: 5Hz). To extract the EMG envelopes, the EMGs were rectified and zero-phase lowpass filtered with a Butterworth filter (half power frequency: 3Hz). The EMG signals were subsequently subsampled to 100Hz. 

The reaching, manipulation, and release phases present in the original dataset were appropriately concatenated. Since previous studies have reported significant correlations between muscle activity and joint accelerations \cite{suzuki2001relationship}, we chose to adopt an acceleration-based kinematic representation. To this aim, we numerically differentiated the hand joint angles twice to compute joint accelerations. All the analyses present in this work were conducted in MATLAB (MATLAB 2020a, The MathWorks, Natick, MA).

\begin{figure*}[ht]
 \center
  \includegraphics[width=\textwidth]{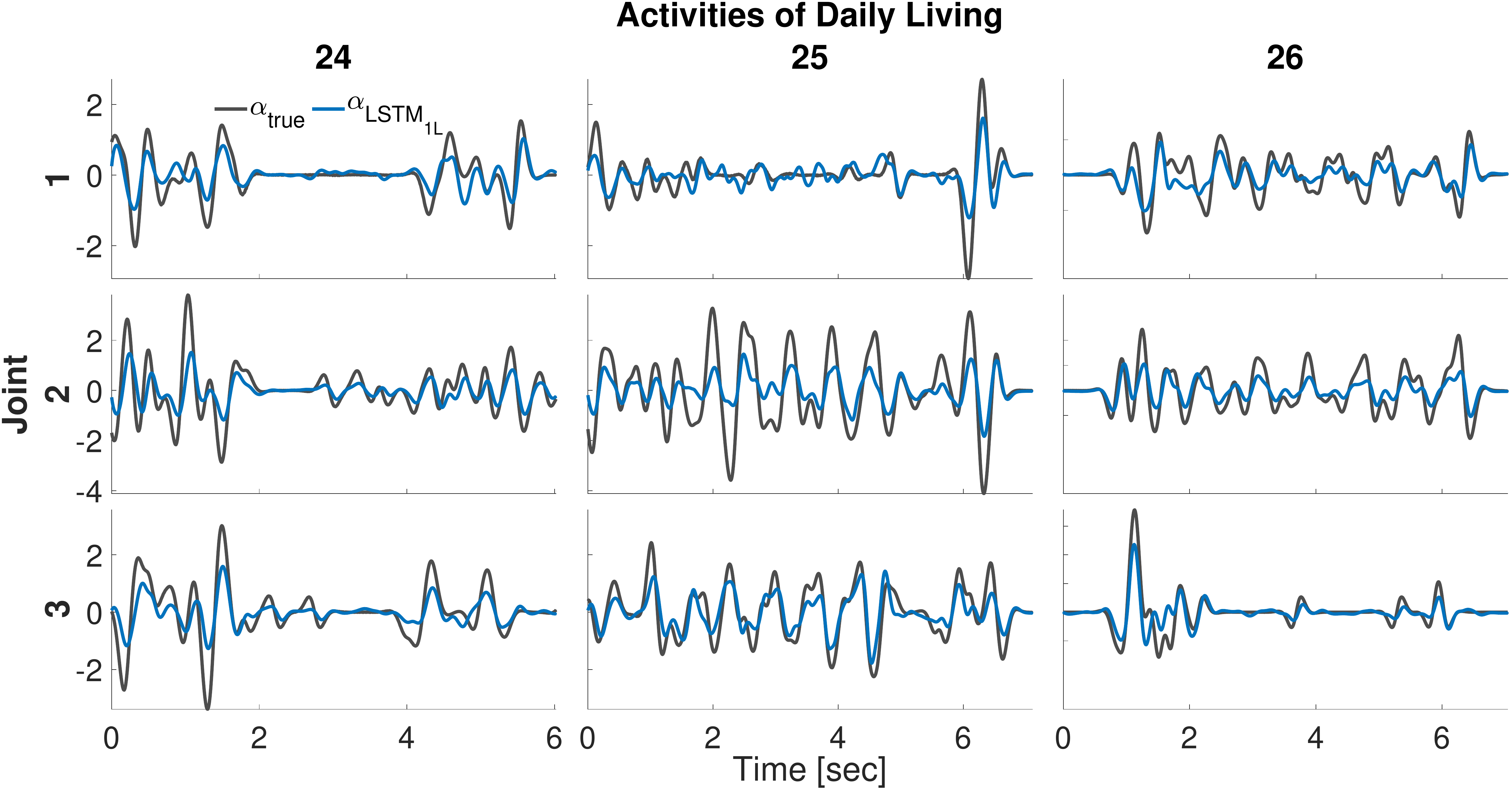}
  \caption{\textbf{Example prediction results.}  Ground-truth (dark gray) and predicted (blue) angular accelerations ($\alpha$) of the three independent activities of daily living (used for testing only), for three (out of 18) representative hand joints, for a representative participant (participant 1). All predictions reported here are made by the 1-layer LSTM network (LSTM$_{1L}$) in the absence of noise.}
    \label{fig2}
\end{figure*}

\subsection{Feature Extraction}
An influential study on feature selection for the classification of EMG signals \cite{phinyomark2012feature} reported that features based on signal energy and those based on frequency content tend to provide superior classification performance. Inspired by this work, we built an EMG feature set composed of EMG envelopes and EMG spectrograms. The spectrograms were computed on 50ms Kaiser windows within the frequency range 20Hz-400Hz; the resulting signals were upsampled in time to 100Hz and binned in frequency (500 bins). \\ 
\indent Subsequently, we applied principal component analysis (PCA) to reduce the dimensionality of the EMG feature set to 25; the extracted projections retained about 99\% of the original data variance.

Since EMG signals might give rise to different movements depending on the starting kinematic properties, such as posture \cite{beringer2020effect}, we extended our feature set by also including information about the previous movements. Specifically, we included the hand joint accelerations observed 30ms in the past. The complete feature set thus includes EMG features and kinematic features. Finally, the resulting features were z-scored. 

\subsection{Data Augmentation}
To avoid over-reliance on the kinematic features, we augmented the training dataset by adding, for each action, 60 additional training examples. In these examples, the kinematic features were corrupted by noise, while the EMG features were left unchanged. Out of the 60 additional examples, 30 were corrupted by white Gaussian noise and 30 by colored noise with a maximum frequency of $10^{-7} {\rm rad} / {\rm sample}$; in both cases the maximum amplitude was $0.1$. Overall, this procedure ensures that the networks learn robust features \cite{zheng2016improving,Ilyas2019AdversarialEA}.
\subsection{Network Architecture}
To learn the mapping between the complete feature set and hand joint accelerations, we used LSTM-based networks \cite{hochreiter1997long}. This choice was motivated by previous work, which showed that the ability of gated recurrent neural networks to learn long-term relationships between signals is instrumental in allowing the prediction of whole-body human motion \cite{martinez2017human,wang2019vred}. Moreover, LSTM-based networks have recently been proved to be effective at predicting wrist position (3 DOFs) \cite{xia2018emg} and wrist flexion/extension (1 DOFs) \cite{kim2020simultaneous} by tracking the EMG activity of 2-5 shoulder and arm muscles.

In brief, LSTM networks are recurrent networks of {\it special} units. Each unit is endowed with two {\it memory cells} (namely, an {\it output} memory ${\bf y}$ and a {\it hidden} memory  ${\bf h}$) and three {\it gates} (namely, an {\it input} gate ${\bf i}$, a {\it forget} gate ${\bf f}$, and an {\it output} gate ${\bf o}$). The hidden memory cell ${\bf h}$ allows the network to store and use important input features indefinitely. The input gate ${\bf i}$ modulates the extent to which incoming information is retained; the forget gate ${\bf f}$ modulates the extent to which old information is erased; finally, the output gate $\bf o$ extracts information from the hidden memory useful to generate an appropriate output ${\bf y}$ for the current input ${\bf u}$. \begin{figure*}[!t]
\center
  \includegraphics[width=\textwidth]{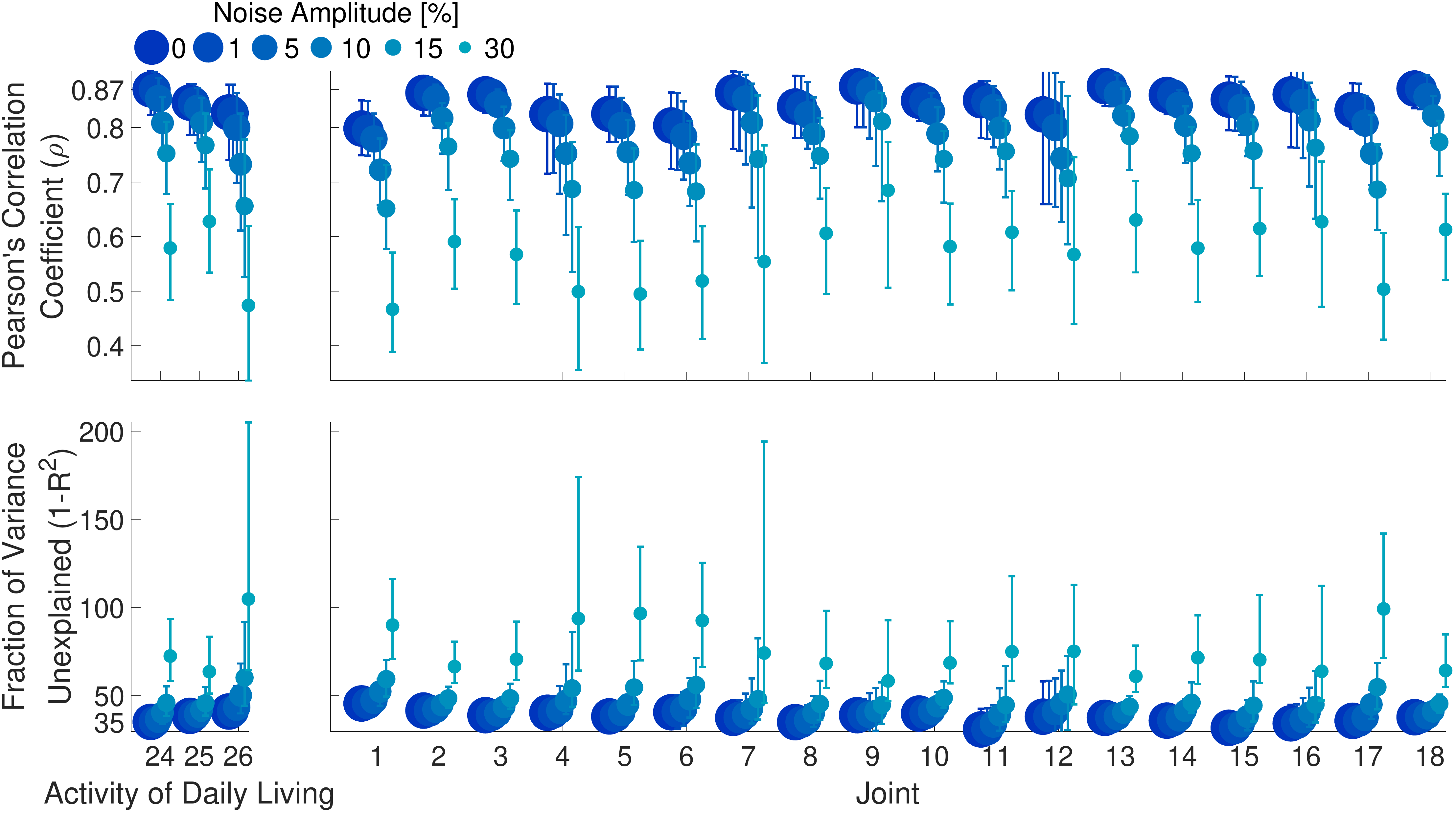}
  \caption{\textbf{Summary prediction results.} (Top): median Pearson's correlations between ground-truth and predicted angular accelerations for the three independent test activities of daily living. (Left): medians across participants, joints, and noise realization, reported as a function of activity of daily living and noise amplitude. (Right): medians across participants, activity of daily living, and noise realization, reported as a function of joint and noise amplitude. (Bottom): corresponding median fractions of variance unexplained. All predictions reported here are made by the 1-layer LSTM network (LSTM$_{1L}$) for increasing amounts of noise. Dot size is proportional to noise amplitude. Error bars represent interquartile ranges.}
    \label{fig3}
\end{figure*} The output, once computed, is stored in the output memory cell. \par
The gating mechanism is defined by the following equations:
\begin{alignat}{3}
{\bf f}^{(t)} &= \sigma({\bf A}_{f} {\bf y}^{(t-1)} &&+ {\bf B}_{f} {\bf u}^{(t)} &&+ {\bf b}_f) \\
{\bf i}^{(t)} &= \sigma({\bf A}_{i} {\bf y}^{(t-1)} &&+ {\bf B}_{i} {\bf u}^{(t)} &&+ {\bf b}_i) \\
{\bf o}^{(t)} &= \sigma({\bf A}_{o} {\bf y}^{(t-1)} &&+ {\bf B}_{o} {\bf u}^{(t)} &&+ {\bf b}_o)
\end{alignat}
where $t$ is the current time step and $\sigma()$ is the element-wise sigmoid function. \par
The hidden memory content is modified according to 
\begin{alignat}{1}
{\bf {\tilde{h}}}^{(t)} &= \tanh({\bf A}_{h} {\bf y}^{(t-1)} + {\bf B}_{h} {\bf u}^{(t)} + {\bf b}_h) \\
{\bf h}^{(t)} &= {\bf f}^{(t)} \odot {\bf h}^{(t-1)} + {\bf i}^{(t)} \odot {\bf \tilde{h}}^{(t)}
\end{alignat}

where  $\odot$ denotes the element-wise product operator.
Finally, the output is determined by 
\begin{equation}
{\bf y}^{(t)} = {\bf o}^{(t)} \odot \tanh({\bf h}^{(t)})    
\end{equation}

\subsection{Hyperparameter Optimization}
In this work, we considered architectures composed of $n_d=\{1,2,3\}$ LSTM layers of $n_h$ units projecting onto a fully connected layer with 36 units, followed by a dropout layer (\cite{srivastava2014dropout}) with probability $0.5$, and a final fully connected layer of 18 units. The dynamics of each LSTM layer are defined by equations (1-6). We chose not to consider deeper LSTM architectures since evidence suggests that the benefit of adding more than two layers is generally limited \cite{karpathy2015visualizing}.

We used Bayesian hyperparameter optimization (bayesopt() —  \cite{snoek2012practical}) to tune $n_h$,  initial learning rate, gradient threshold, and L2 regularization strength. The search ranges were $[30,600]$, $[10^{-4}, 10^{-1}]$, $[.2, 1]$ and $[10^{-10}, 10^2]$, respectively. The available LSTM units were equally distributed among the $n_d$ layers. The selected cost function was the fraction of unexplained variance on the validation set.

Hyperparameter optimization automatically stopped when the maximum number of 100 cost function evaluations was reached. To efficiently evaluate a candidate hyperparameter set, we trained the parameters of the corresponding network for only three epochs. To further speed up the training time, we based hyperparameter optimization only on the data recorded from participant 1. Following this procedure, we selected, for each depth $n_d$, the three best networks in terms of validation error. Such networks were further trained for 25 epochs with subject-specific data only. Details about the parameter optimization are provided in the following section. Overall, this strategy allowed us to retrieve a good hyperpameter set reasonably fast: for example, for $n_d=1$, hypeparameter optimization was completed in less than one hour (i.e., $54' 47''$).
\begin{figure*}[!t]
\center
  \includegraphics[width=\textwidth]{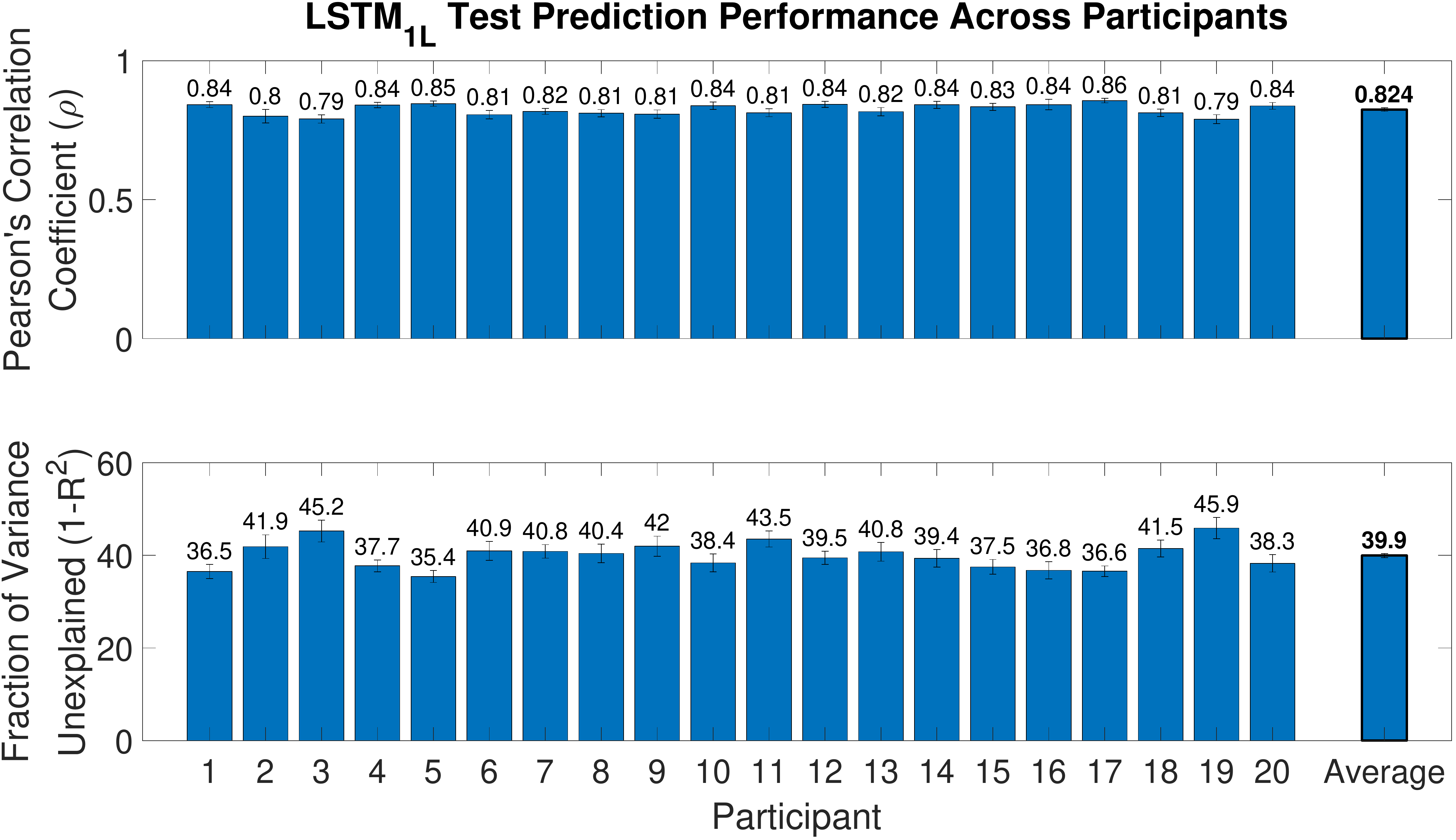}
  \caption{\textbf{Summary performance across subjects.} (Top): Pearson's correlations between ground-truth and LSTM$_{1L}$-predicted angular accelerations for the three independent test activities of daily living. Data are averaged across joints and tasks, and reported as a function of participant. The rightmost bar reports the average across participants. Error bars represent standard errors.
  (Bottom): corresponding fractions of variance unexplained. All predictions reported here are made by the 1-layer LSTM network (LSTM$_{1L}$) in absence of noise.}
    \label{fig4}
\end{figure*}
\subsection{Parameter Training}
The networks were implemented and trained in MATLAB, using the Neural Network Toolbox. Parameter optimization was performed using adaptive moment estimation (Adam, \cite{kingma2014adam}) with a piecewise learning rate schedule with a drop period of 10 epochs and a drop factor of 10\%. To speed up training, we performed all the optimizations on a shared GPU (NVIDIA GeForce RTX 2070) using mini batches of 40 sequences. This configuration allowed us to fully train a network in a limited amount of time: for example, for $n_d=1$, the training speed was $8.25$ sec/epoch. Parameter optimization automatically stopped if the validation error did not decrease for three consecutive training epochs, or if the maximum number of 30 epochs was reached. For training, validation, and test we used the EMG signals and corresponding hand kinematics recorded during the execution of the tasks 1-20, 21-23, and 24-26, respectively. We refer to Table \ref{tab:my-table} for further details about the selected tasks.

\subsection{Assessment of Network Performance}
To measure the reconstruction quality of the network's predictions, we computed the Pearson's correlation coefficients ($\rho$) between ground-truth and predicted angular accelerations, and the normalized reconstruction errors ($\epsilon$). The normalized reconstruction error is defined as the fraction of unexplained variance (i.e., $\epsilon=1-R^2$).

To assess the potential benefit of the recurrence present in LSTM networks, we also considered standard non-linear feedforward (FF) networks as a baseline. Specifically, we considered FF networks of depth $n_d=\{1,3,10\}$ with hyperbolic tangent activation function. A dropout layer with probability $0.5$ was added between each couple of layers. For each depth, similarly to what was done for LSTM networks, we performed Bayesian hyper-parameter optimization. In this case, we set the maximum number of units to 3000. All the other parameter ranges are as above.

To measure the robustness of the trained networks, we also assessed their ability to make accurate predictions in the presence of random noise corrupting the angular acceleration measurements. Specifically, we considered noise levels of amplitudes $\xi_a = \{0, 1, 5, 10, 15, 30\}\%$. The amplitudes are defined as percentages of the maximum angular acceleration measured during a task, across all joints. 

In the following sections, we report the test performance (measured on the independent activities of daily living 24, 25, and 26) of the resulting six networks: three LSTM and three FF networks.

%% file: Results.tex
Figure \ref{fig1} summarizes the performance of the networks trained after hyper-parameter optimization: for low levels noise ($\xi_a\leq5\%$) most networks are able to predict the joint angular velocities measured during the performance of the test tasks well ($\rho>.7$). This is not trivial since these test tasks were not used at any stage of the training. The only exception is represented by the 10-layer feedforward network, which shows poor performance also for low levels of noise. In this low noise regime, the 1-layer and 3-layer feedforward networks (FF$_{1L}$, FF$_{3L}$) and the 1-layer LSTM network (LSTM$_{1L}$) display the best performance ($\rho>.8$).

However, as the noise increases ($\xi_a>5\%$ —  simulating a more realistic scenario of online usage of the prosthesis), the performance pattern changes drastically: whereas all feedforward networks become unable to make accurate predictions ($\rho<.5$), all LSTM-based networks are still able to track the ground-truth data with sufficient accuracy ($\rho>.6$). Overall, the network LSTM$_{1L}$ displayed the most accurate and robust performance across all tested noise levels (Fig. \ref{fig1} — right), and will be the main focus of the following sections.

Figure \ref{fig2} shows ground-truth and LSTM$_{1L}$ predictions of angular accelerations during test tasks for few representative joint angles. It is worth noting that the network makes good predictions even though the hand kinematics vary substantially across tasks. As a matter of fact, despite the heterogeneity of hand kinematics across joints and the variability across tasks, LSTM$_{1L}$ average performance does not drastically change across joints (Fig. \ref{fig3} — right) or task (Fig. \ref{fig3} — left). However, it is clear that the network's prediction performance for some joints (e.g., joints 1, 4, and 17) degrades substantially when the noise amplitude is particularly high.

Overall LSTM$_{1L}$ displayed satisfying average prediction performance ($\rho=.824$, $\epsilon=39.9\%$ — Fig. \ref{fig4}). Moreover, despite the high across-subjects behavioral variably, consequence of the loose task constraints, the prediction performance was remarkably stable across subjects ($\rho\geq.79$, $\epsilon\leq45.9\%$ — Fig. \ref{fig4}).

%% file: Conclusion.tex
In this work, we trained LSTM and feedforward (FF) neural networks to predict hand kinematics during the execution of activities of daily living from forearm EMG muscles. In the presence of low noise levels, shallow FF networks and LTSM-based networks tend to perform similarly. In the presence of higher noise levels — which simulate a scenario that is more relevant for the online control of a hand prosthesis — LSTM-based networks tend to over-perform FF networks. This suggests that the mapping learned by the recurrent models is more robust and more suitable for online prosthesis control. Stacking multiple LSTM layers did not provide any apparent additional benefits, in agreement with previous reports \cite{karpathy2015visualizing}. In fact, we found that the best performing network was the one with only one hidden LSTM layer: LSTM$_{1L}$

Remarkably, we showed that the mapping learned by LSTM$_{1L}$ is robust enough to generalize across activities of daily living it was never trained on. Furthermore, the overall performance is stable across tasks and hand joints. To the best of our knowledge, these are the first prediction results to ever be reported in the literature on this dataset.

Notable technological advances in mechatronics have led to the development of highly flexible human-like motorized hand prostheses. Nevertheless, the complex functional mapping between forearm muscle signals and hand kinematics, together with the tendency to validate control algorithms on datasets of highly simplified movements, have severely hampered the design of efficient algorithms capable of providing natural and dexterous control of these devices. In this work, we introduced a method based on LSTM networks to learn such a mapping. Critically, unlike previous works, we validated the method on a dataset of realistic arm and hand movements involving complex interactions with daily living objects. Our results show that the mapping is stable across postures and interaction types. This suggests that our method can be used to provide a natural interface between forearm muscle activity and state-of-the-art motorized hand prostheses. Future work will deal with developing a suitable interface layer between the network's predictions and the actuators of the prosthesis.